\documentclass[letterpaper, 10 pt, conference]{ieeeconf}  

\IEEEoverridecommandlockouts                              

\usepackage{amsmath} 
\usepackage[english]{babel}
\usepackage[utf8x]{inputenc}
\usepackage{graphicx}
\usepackage{url} 
\usepackage{cite}
\usepackage{subcaption}
\usepackage{siunitx}
\usepackage{multirow}
\usepackage{booktabs}
\usepackage[nolist]{acronym}
\usepackage{algorithm}
\usepackage{algpseudocode}
\usepackage{amsfonts}

\newcommand{\ik}{\ac{IK}}
\newcommand{\fk}{\ac{FK}}
\newcommand{\ann}{\ac{ANN}}
\newcommand{\dof}{\ac{DOF}}
\newcommand{\gnn}{\ac{GNN}}
\newcommand{\gp}{\ac{GP}}
\newcommand{\sr}{\ac{SR}}
\newcommand{\mae}{\ac{MAE}}

\begin{document}

\title{\LARGE \bf
The Road to Learning Explainable Inverse Kinematic Models: Graph Neural Networks as Inductive Bias for Symbolic Regression
\thanks{This work was supported by the German Research Foundation (DFG) under project grants SPP2331 and SPP2364.}}

\author{Pravin Pandey$^{1}$, Julia Reuter$^{1}$, Christoph Steup$^{1}$ and Sanaz Mostaghim$^{1,2}$
\thanks{$^{1}$All authors are with Faculty of Computer Science, Otto-von-Guericke-University Magdeburg, Germany {\tt\small \{pravin.pandey, julia.reuter, steup, sanaz.mostaghim\}@ovgu.de}}%
\thanks{$^{2}$Sanaz Mostaghim is with Fraunhofer Institute for Transportation and Infrastructure Systems IVI, Dresden, Germany%
}}

\begin{acronym}
\acro{IK}{Inverse Kinematics}
\acro{FK}{Forward Kinematics}
\acro{GP}{Genetic Programming}
\acro{ANN}{Artificial Neural Network}
\acro{GNN}{Graph Neural Network}
\acro{DOF}{Degree of Freedom}
\acro{SR}{Symbolic Regression}
\acro{MAE}{Mean Absolute Error}
\end{acronym}

\maketitle
\thispagestyle{empty}
\pagestyle{empty}

\begin{abstract}

This paper shows how a Graph Neural Network (GNN) can be used to learn an Inverse Kinematics (IK) based on an automatically generated dataset. The generated \acl{IK} is generalized to a family of manipulators with the same \acf{DOF}, but varying link length configurations. The results indicate a position error of less than 1.0\,cm for 3 \acs{DOF} and 4.5\,cm for 5~\acs{DOF}, and orientation error of 2$\mathbf{^\circ}$ for 3~\acs{DOF} and 8.2$\mathbf{^\circ}$ for 6~\acs{DOF}, which allows the deployment to certain real-world problems. However, out-of-domain errors and lack of extrapolation can be observed in the resulting \acs{GNN}. 
An extensive analysis of these errors indicates potential for enhancement in the future. 
Consequently, the generated GNNs are tailored to be used in future work as an inductive bias to generate analytical equations through symbolic regression.

\end{abstract}



\section{Introduction}

Nowadays, robotic manipulators become more and more integrated in applications
as the price decreases and the capabilities increases, some examples of this
process are the widespread use of manipulators in multiple RoboCup leagues like
the @Home or the @Work ones\cite{robocup}. Another large use case is automation
and support of scientific experiments in various
environments~\cite{space_manipulators}, which involves small-scale automation
with a strong variation of objects. Because of these new applications, many new
manipulator types are entering the market, vastly expanding the range of kinematics, size and
\dof{}. However, the creation of \ik{}
for newly designed manipulators may be challenging if it does not follow the
classical constraints of having 6 \dof{} and a typical joint configuration.
Soft~robotic~manipulators~\cite{soft_robotic_manipulators} further enhance the
problem of \ik{} because they provide many new applications, but need
entirely new planning and control strategies.

Using a learned \ik{} instead of manually designed ones may simplify the design
of new types of manipulators because the tedious and complex task of developing
an analytical \ik{} could be automized. It may even enable a smooth transition
between classical rigid manipulators and soft manipulators, which may be handled
as rigid manipulators with a dynamically changing joint configuration. One large
drawback of the typical learning methods like an \ann, is the lack of
explainability and also trustworthiness, especially in out-of-domain cases. For
this purpose, we envision a \gp-based system to learn the \ik{}, which would
provide an analytical solution enabling explanation and verification. However,
the training of a \gp-based \ik{} is challenging and often not
successful~\cite{chapelleClosedFormInverse2001}. Consequently, we propose a
\gnn-based learning system to generate a learned \ik{} as a stepping stone for a
\gp-based \ik{}. We believe the \gnn-based \ik{} to provide the necessary
inductive bias because of their internal structure, which enables a
decomposition of the complex problem into smaller
sub-problems~\cite{cranmerDiscoveringSymbolicModels2020}.
This simplifies the
training of \gp{} to enhance convergence, especially for complicated
manipulators. Out of the two steps involved in this approach, this paper
addresses the first step and examines the performance of the \gnn-based learned
\ik{} to assess its applicability towards a \gp-based \ik{}.


\section{Related Work}

\subsection{GNN for Inverse Kinematics}
\ann s are a relatively well-understood approach to solve the \ik{} problem for robotic manipulators, as a plethora of related studies suggest~\cite{aristidouInverseKinematicsTechniques2018,dembysStudySolvingInverse2019,wagaaAnalyticalDeepLearning2023,el-sherbinyComparativeStudySoft2018}. 
\gnn{}s are a type of \ann{} that operate on graph-like structures using messages
passed between nodes connected by edges. Currently, few publications are
available relating to robotic manipulators with \gnn{}s. 
The first study by Sanchez~et~al.~\cite{sanchez1} proposed using \gnn{}s to learn the behaviour of complex physical systems like manipulators. Their approach was able to plan a trajectory the manipulator had to follow. However, error values in meters are not reported, which hampers comparison. In a later work~\cite{sanchez2learning}, they proposed an
encoder-decoder architecture to overcome the limitation of short trajectories of
the first work. 
This method has not yet been tested on manipulators. 
Kim et al.~\cite{kim} used a \gnn{} architecture to learn robotic manipulators' forward and inverse kinematics. The approach is tested on simulated 2D
manipulators, which again is difficult to compare to our 3D workspace \gnn{}.
 
\subsection{Symbolic Models for Inverse Kinematics} \sr{} is the identification
of mathematical equations from data, which can help to gain a more profound
understanding of the problem and overcome the black-box nature of \ann{}s. 
\gp{} is a popular method for \sr{} based on evolutionary algorithms, which has recently gained importance for scientific equation discovery. 
Two early publications by Chapelle~et~al.~\cite{chapelleClosedFormInverse2001,chapelleClosedFormSolutions2004} use \gp{} to discover \ik{} equations for 6 \dof{} manipulators. 
In both papers, the loss function is the rooted mean square error of the joint angles penalized by the length of the equation. 
The results were unusable for real-world applications because of a lack of accuracy. 
Recently, using \gnn{} as an inductive bias for \gp{} showed promising results, with examples including the recovery of Hamiltonian and Newtonian dynamics~\cite{cranmerDiscoveringSymbolicModels2020}, collective behaviour of swarm robotics~\cite{powersExtractingSymbolicModels2022}, modelling hydrodynamic forces in particle-fluid flows~\cite{reuter_graph_2023} as well as Kepler's orbital mechanics of the solar system~\cite{lemosRediscoveringOrbitalMechanics2023}. Consequently, we propose the use of \gnn{}s to learn the \ik{} problem and, in a later step, replace the internal parts of the network with simpler analytic models generated by \sr.


\section{GraphMatic}
Many systems in the engineering and physics domain can be modelled as a graph, containing objects and their interactions.
\gnn{}s  are a type of deep neural networks that operate on graph-structured data~\cite{bronsteinGeometricDeepLearning2017}. 
Their separable internal structure motivates the use of GNNs as an intermediate step to learn mathematical equations from high-dimensional data through symbolic regression, as it allows breaking down complex problems into smaller sub-problems which are tractable for SR algorithms~\cite{cranmerDiscoveringSymbolicModels2020}. 
In the following, we introduce message passing networks and describe the GraphMatic approach to perform this intermediate step and learn an inverse kinematics model of arbitrary robotic manipulators.

\subsection{Inverse Kinematics Problem}
\label{sec:problem}
The general structure of a robotic manipulator is an open kinematic chain with multiple, typically rotational joints, which are connected by links. 
Non-rotational joints, such as prismatic joints, are left out of consideration in this paper. 
The joint rotation parameters $\boldsymbol{\theta} = [\theta_1, \dots,\theta_{DOF}]$ are the variables of interest for the \ik{} problem. 
A robotic manipulator operates in a six-dimensional environment $\mathbf{X} = [\mathbf{p}, \mathbf{o}]^T = [x,y,z,\Phi, \Theta, \Psi]^T$. 
In this pose representation, $\mathbf{p} = [x, y, z]^T$ indicates the position of the end-effector in Cartesian space, and $\mathbf{o}=[\Phi, \Theta, \Psi]^T$ denotes the end-effector orientation in Euler angles.
The goal of the \fk{}, denoted by $g$, is to determine the pose of the end-effector given the robot joint angles, i.e.,\ $\mathbf{X} = g(\boldsymbol{\theta})$. 
The \ik{} problem aims at finding the robot joint angles according to a target pose, i.e.,\ $\boldsymbol{\theta} = g^{-1}(\mathbf{X})$. 
In contrast to the \fk{}, there are multiple solutions to the \ik{} problem, which makes the problem considerably more complex.
In general, finding a closed form solution is of great interest, i.e., finding an explicit relationship between a joint variable $\theta_i$ and the elements of the six-dimensional pose vector $\mathbf{X}$.

In this paper, we look at the \ik{} problem from two perspectives: a \textit{direct estimation}~(DE) of the joint angles given the target pose, and a \textit{reference guided}~(RG) approach inspired by robot path planning applications. 
In any path planning application, the route to reach the final position is accomplished in multiple intermediate positions with small increments. 
Following the same strategy in the RG architecture, we provide a reference angle $\boldsymbol{\tilde{\theta}}$, which is close to the actual joint angle.
In other words, IK-DE computes the target joint configuration only from the pose vector, i.e., $\boldsymbol{\theta} = g^{-1}(\mathbf{X})$.
IK-RG additionally receives a reference angle close to the target angle as an input variable, so that $\boldsymbol{\theta} = g^{-1}(\mathbf{X,\boldsymbol{\tilde{\theta}}})$.
It should be noted that our assumption is that close intermediate positions also have close joint angles, although this is not necessarily always the case.

\subsection{General Approach}
The GraphMatic approach represents a robotic manipulator as a graph $g = (V, E)$, consisting of a set of nodes $V$ and edges $E$.
Here, $V = \{\mathbf{v}_i\}_{i=1:N^v}$ denotes the nodes, which in our method correspond to the joints of the robotic manipulator.
The total number of nodes in the graph is denoted by $N^v$.
The node features, $\mathbf{v}_i$, describe characteristics such as joint angle values, joint types, or angular offsets.
In our setting, each joint or node contains information about the target pose of the end effector.
The edges, $E = \left\{ \left(\mathbf{e}_k, r_k, s_k\right) \right\}_{k=1:N^e}$, represent the connections between these joints.
$N^e$ indicates the total number of edges in the graph.
The edge feature vector, $\mathbf{e}_k$, includes information such as link length or translational offset.
    
\textit{Message Passing Neural Networks}~(MPNN) are a type of GNN that operate on graph structures by iteratively passing and aggregating information (or “messages”) between nodes in a graph.
This approach allows each node to incorporate local structural information and features from its surrounding nodes, enabling the network to learn complex dependencies and patterns within the graph. 
It does that by performing consecutive steps of message computation, aggregation, and node update, as outlined in the following equations:
\begin{align}
    \mathbf{m}_{i,j} &= \phi^e(\mathbf{v}_i, \mathbf{v}_j, \mathbf{e}_{i,j}) &\text{\textit{message computation }} \label{eq:msg_comput}\\ 
    \mathbf{m}_j &= \rho^{e \rightarrow v}(\{ \mathbf{m}_{i, j}\}_{i \in \mathcal{N}(j)} ) &\text{\textit{message aggregation }} \label{eq:msg_agg}\\
    \mathbf{v}_j' &= \phi^v(\mathbf{v}_j, \mathbf{m}_j) &\text{\textit{node update }} \label{eq:node_update}
\end{align}

The node feature vectors $\mathbf{v}_i$ and $\mathbf{v}_j$ of nodes $i$ and $j$, as well as the edge feature vector $\mathbf{e}_{i,j}$ of the edge between nodes $i$ and $j$ are the input to the edge model $\phi^e$. 
This model computes a message vector $\mathbf{m}_{i,j}$, which is sent from node $i$ to node $j$ in Eq.~\ref{eq:msg_comput}.
The incoming messages of a node $j$, sent from nodes in its neighbourhood $\mathcal{N}(j)$, are aggregated in Eq.~\ref{eq:msg_agg} using the aggregation function $\rho^{e \rightarrow v}$.
In MPNNs, the elementwise summation is usually applied so that $\rho^{e \rightarrow v}=\sum_{i \in \mathcal{N}(j)}{\mathbf{m}_{i,j}}$.
The aggregated message vector $\mathbf{m}_j$, together with the current node state $\mathbf{v}_j$, is then input to the node model $\phi^v$ to compute the updated node $\mathbf{v}_j'$ (Eq.~\ref{eq:node_update}).
This updated node contains the actual target variable of each joint, i.e., the value of the joint angle required to reach the target pose.
The formal definition of MPNNs also includes the computation of a global property for the entire graph\cite{battagliaRelationalInductiveBiases2018}.
For many systems, including the one addressed in this paper, computations on the edge and node level suffice, i.e., the update step is complete once Eq.~\ref{eq:node_update} is executed.
Both $\phi^{e}$ and $\phi^{n}$ use shared parameters to compute messages and update nodes.
In other words, it assumes that the first and second joint interact in the same way as the fourth and fifth.

\begin{figure}
    \centering
    \includegraphics[width=0.65\linewidth]{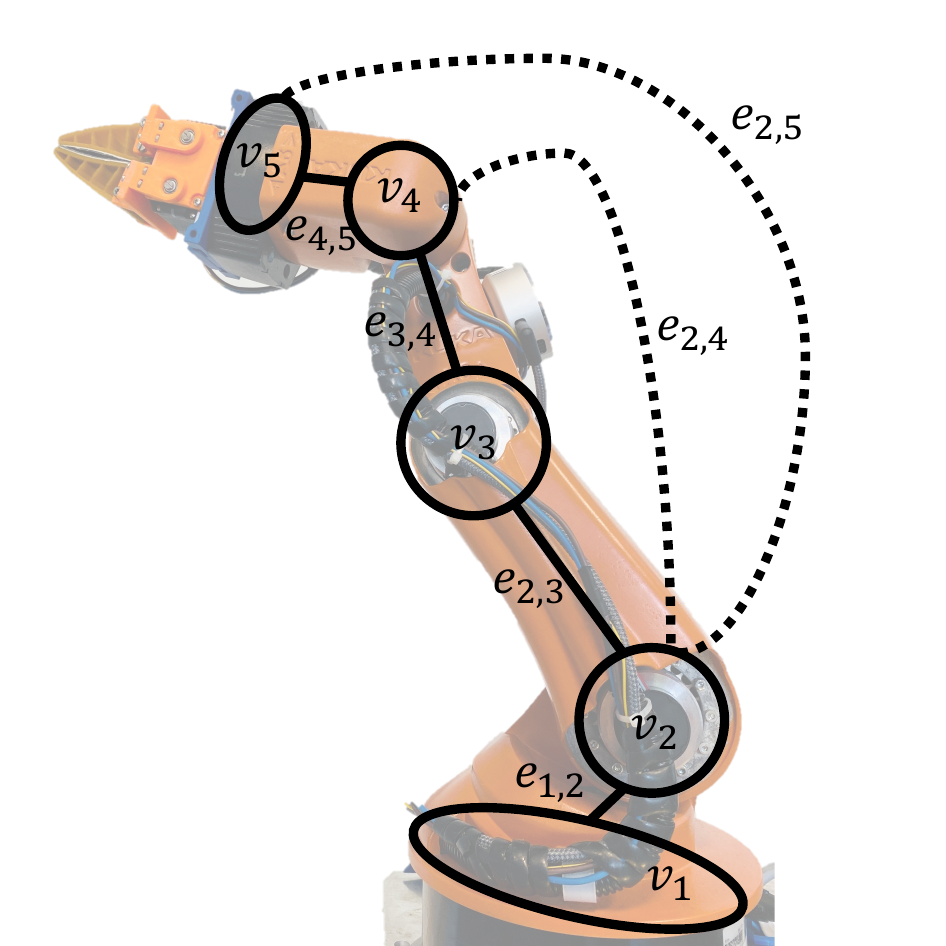}
    \caption{Graph projected on a manipulator with 5~DOF. Nodes $v_i$ and $v_j$ are connected by an edge $e_{i,j}$. Solid lines indicate neighbourly connections between nodes. Dashed lines indicate the additional edges for full connectivity, using node $v_2$ as an example to avoid visual clutter. More edges are added for all other nodes in the same way.}
    \label{fig:enter-label}
\end{figure}

\subsection{Variants of GraphMatic}
\label{sec:variantsGraphMatic}
As this is the initial application of this approach to arbitrary robotic manipulators, we propose multiple variants of GraphMatic to experiment with its characteristics and assess its effectiveness.
First, the impact of \textit{different underlying connection types} between the joints can be evaluated.
A \textit{neighbourly connection} between joints is inherent to robotic manipulators, where each joint in the kinematic chain is only connected to the previous and subsequent joint through links.
This variant relies on local connectivity to learn an \ik{} model, with limited information about the area outside the immediately neighbouring joints.
We furthermore propose a \textit{full connection} variant, where each joint receives messages from all other joints, to assess how the local connectivity influences the model performance compared to a global connectivity.
Second, we evaluate our approach on \textit{two modes of \ik{} problems} that appear in real-world applications, as described in Sec.~\ref{sec:problem}.
This makes a total of four variants: First, GraphMatic-DE-N and GraphMatic-DE-F solving the \ik{} with a direct estimation approach defined as $\boldsymbol{\theta} = g^{-1}(\mathbf{X})$ using neighbouring (N) and full (F) graph connectivity, i.e., $\boldsymbol{\theta} = g^{-1}(\mathbf{X})$.
Secondly, GraphMatic-RG-N and GraphMatic-RG-F with an additional reference joint angle in the set of node features $\boldsymbol{\theta} = g^{-1}(\mathbf{X,\boldsymbol{\tilde{\theta}}})$. 


\section{Experimental Setup}

We investigate the viability of the four algorithm variants proposed in Sec.~\ref{sec:variantsGraphMatic}.
To this end, we perform experiments for different manipulators with 3, 5, and 6 \dof{}.
Multiple network sizes are assessed on the 3~\dof{} dataset, of which the best performing will serve as a setting for higher DOF.

\subsection{Data Generation}
\label{sec:datasets}
To learn the \ik{} problem using GraphMatic, we generated the data using our Python-based framework. 
The framework generates collision-free, user-specified link lengths denoted by $U$ and dataset length for 3, 5 and 6 \dof{} manipulators. 
The end-effector pose is calculated using the D-H convention. 
For new instances, a vector $\boldsymbol{\theta}_{new}$ is created, with \dof{} random values drawn from a uniform distribution within the movement range of each joint.
To cover the workspace well, we implement a mechanism that guarantees new samples are distinct from existing ones in the dataset.
If the condition 
\begin{align}
\min_{\overline{\boldsymbol{\theta}} \in \text{Dataset}} \lVert \overline{\boldsymbol{\theta}} - \boldsymbol{\theta}_{\text{new}} \rVert_{\infty} > 1^\circ 
\end{align}
is satisfied, $\boldsymbol{\theta}_{\textrm{new}}$ is accepted and else rejected. 
As well as, if the combination of joint angles $\boldsymbol{\theta}_{\textrm{new}}$ leads to self-collisions or collision with the ground, the combination of joint angles is still rejected. 
The step is repeated until the dataset length equals the length specified by the user. 

\begin{table}[t]
    \centering
    \scriptsize
    \caption{Configuration for 3 \dof{} manipulator}
    \label{tab:3DoF config}
    \begin{tabular}{llllll}
        \hline
         Joint & $\theta$ $[^{\circ}]$ & $\theta_{\textrm{off}}$ $[^{\circ}]$ & $a$ [cm] & $d$ [cm] & $\alpha$ $[^{\circ}]$   \\
         \hline
         1 & 0-360  & 0 & 0 & 40-60 & 90  \\
         2 & 0-360 & 90 & 24-36 & 0 & 0   \\
         3 & 0-360 & 0 & 15-20 & 0 & 0   \\
         \hline
    \end{tabular}
\end{table}
\begin{table}[t]
    \centering
    \scriptsize
    \caption{Configuration for 5 \dof{} manipulator}
    \label{tab:5DoF config}
    \begin{tabular}{llllll}
        \hline
         Joint & $\theta$ $[^{\circ}]$ & $\theta_{\textrm{off}}$ $[^{\circ}]$ & $a$ [cm] & $d$ [cm] & $\alpha$ $[^{\circ}]$   \\
         \hline
         1 & 0-360  & 0 & 0 & 40-60 & 90  \\
         2 & 0-360 & 90 & 24-36 & 0 & 0   \\
         3 & 0-360 & 0 & 15-20 & 0 & 0   \\
         4 & 0-360 & -90 & 9-13 & 0 & -90   \\
         5 & 0-360 & 0 & 0 & 5-8 & 0   \\
         \hline
    \end{tabular}
\end{table}
\begin{table}[t!]
    \centering
    \scriptsize
    \caption{Configuration for 6 \dof{} manipulator}
    \label{tab:6DoF config}
    \begin{tabular}{llllll}
        \hline
         Joint & $\theta$ $[^{\circ}]$ & $\theta_{\textrm{off}}$ $[^{\circ}]$ & $a$ [cm] & $d$ [cm] & $\alpha$ $[^{\circ}]$   \\
         \hline
         1 & 0-360  & 0 & 0 & 40-60 & 90  \\
         2 & 0-360 & 90 & 24-36 & 0 & 0   \\
         3 & 0-360 & -90 & 15-20 & 0 & -90   \\
         4 & 0-360 & 0 & 0 & 9-13 & 90   \\
         5 & 0-360 & 0 & 5-8 & 0 & -90   \\
         6 & 0-360 & 0 & 0 & 5 & 0   \\
         \hline
    \end{tabular}
\end{table}

The first link $U_\textrm{1}$ takes on values between 40-60 cm, and the remaining link lengths are calculated using the formula
\begin{equation}
    U_{i+1} = \max\{0.75 U_i \pm 0.15 U_{i}; \; 5\textrm{cm}\}
\end{equation}
to guarantee a minimum link length of 5~cm.  
Tab.~\ref{tab:3DoF config},~\ref{tab:5DoF config}, and~\ref{tab:6DoF config} show the ranges for all the D-H parameters for 3, 5, and 6 \dof{} manipulators, respectively. 
The $\alpha$ value is fixed for all the joints, and the $\theta$ value is randomly generated using a uniform distribution between [0,360). 
The offset values~$\theta^{\circ}_{\textrm{off}}$ are chosen so that the manipulators take on a specific neutral position when all the joint values are equal to $0$. 
The final dataset contains the features $a_\textrm{{i}},d_\textrm{{i}},\alpha_\textrm{{i}},\theta_\textrm{{i}}, \theta_{\textrm{i,off}}, A_\textrm{{i}}, x, y, z, \Phi, \Theta, \Psi$. 
$A_{i}$ represents the D-H matrix for each joint, and $i$ denotes the joint value. 
The $\theta_\textrm{{i}}$ is available both in degrees and radians.

In this work, we generate ten different link length configurations for each \dof{} using our framework. 
We create 0.5, 1.0, and 2.0 Million data points for 3, 5, and 6 \dof{} manipulators for each configuration, summing up to 5, 10 and 20 Million data points, respectively.
Nine configurations are used for training, and the configuration with the overall longest kinematic chain is employed for testing the extrapolation capabilities of the learned models to predict samples farther than the previously seen reachable area. 
Additionally, we create validation datasets with 10,000 samples for every \dof{} using link length combinations unused for training, that are within the reachable area of the training datasets, i.e., no extrapolation is required by the \gnn{}. 

\subsection{Algorithm Configuration}

The training data is split in an 80\%-20\% ratio for training and early stopping detection, which stops the algorithm's learning once the loss on the early stopping set increases over a few defined epochs. 
We perform an interactive selection of network parameters iteratively, starting with 3 \dof{}, aiming to finding the simplest network structure without using regularization like dropout.
GraphMatic for 3 \dof{} is trained with two hidden layers along with 22, 32, and 42 neurons in each layer, similar to the work of~\cite{rourkela}.
Based on the results of 3 \dof{}, we select 35 neurons per hidden layer for 5 and 6 \dof{}, and provide a deeper network with 4 layers to enable the learning of more complex features for the more complex problems. 
We performed preliminary experiments with varying batch sizes and got satisfactory results with a batch size of 5000, which was superior to all other batch sizes. 
Each network configuration is trained once, and evaluated on numerous test configurations.

\begin{table}[t]
    \centering
    \scriptsize
    \caption{Parameter settings used for GraphMatic}
    \begin{tabular}{ll}
        \hline
        Parameters & Values \\
        \hline
        Nonlinearity &  Rectified Linear Unit (ReLU) \\
        No. hidden layers $l$ & 2, 4\\
        No. neurons per layer $n$ & 22, 32, 42, 35\\
        Max. learning rate & 0.002 \\
        Batch size & 5000 \\
        Optimizer & AdamW \\
        Loss function & Mean Squared Error (MSE)\\
        Training duration & 1000 epochs\\
        Early Stopping & 10 epochs\\
        Train/early stop data ratio & 0.8 / 0.2, random split \\
        Size of message vector & 6 \\
        Size of node vector &  1 (target variable $\theta_i$)\\
        No. run & 1 \\
        Node features & $x$, $y$, $z$, $\Phi$, $\Theta$, $\Psi$, $\theta_{i,\textrm{off}}$, ($\theta_{i,\textrm{ref}}$ for RG)\\
        Edge features & $\alpha_{i}$, $l_{i}$\\
        \hline
    \end{tabular}
    \label{tab:my_label}
\end{table}


\section{Evaluation}

We use the $R^{2}$ metric to evaluate the variants of GraphMatic, which measures the proportion of the variance in the dependent variable that is explained by the independent variables~\cite{mae_r2}. 
The best value for $R^{2}$ is 1; the worst is $-\infty$. 
Generally, GraphMatic is trained and evaluated on the joint angles $\boldsymbol{\theta}$.
In post-processing, we calculate the predicted pose $\hat{\mathbf{X}}$ using the \fk{} equations and the predicted joint angles $\hat{\boldsymbol{\theta}}$, and examine the position and orientation errors. 
We employ the Euclidean distance between the predicted and target position in three dimensions as the position error. 
Since orientation has circular behaviour, we compute the distance between two Euler angles as the convex angle, i.e., the smallest angle between them when measured on the unit circle.
For the orientation error, we compute the \mae{} of the three Euler angles.
We evaluate the learned models exclusively on the test dataset and do not perform experiments on real manipulators, as such experiments are unlikely to provide additional insights into the given problem because the positioning error of real-world manipulators is typically lower than the error produced by the \ik{}.

For brevity, each \gnn-configuration is named ${\{\textrm{DE,RG}\}\textrm{-}\{\textrm{N,F}\}\textrm{-\dof{}-}l\textrm{-}n}$, where DE and RG specify the GraphMatic approach. N and F specify neighbouring or full connection. $l$ and $n$ follow the definition of Tab.~\ref{tab:my_label}.

\subsection{GraphMatic-DE}
Tab.~\ref{learning_overview} displays how well different configurations of GraphMatic predict the target joint angles $\boldsymbol{\theta}$ using the $R^2$ metric. 
The GraphMatic-DE for 3 \dof{} performed better with 32 neurons compared to 22 neurons, and further increasing the number of neurons to 42 in each layer did not improve the result. 
For 5 and 6 \dof{} manipulators, we observe that the GraphMatic-DE was unable to learn an \ik{} model with $R^{2}$ close to 0, indicating that the model is not better than always predicting the mean of the target variable.
We summarize that a direct estimation of the required joint angles $\boldsymbol{\theta}$ solely from the target pose is possible for simple 3 \dof{} manipulators but impossible for more complex ones.

\begin{table}[t]
\caption{Results of GraphMatic with mean loss and $R^2$, evaluated on the joint angles $\boldsymbol{\theta}$.}\label{learning_overview}
\centering
\scriptsize
\begin{tabular}{lllll}\toprule
GraphMatic & \multicolumn{2}{c}{Training} & \multicolumn{2}{c}{Test}
\\\cmidrule(ll){2-3}\cmidrule(ll){4-5} 
& {Loss} & {$R^{2}$} & {Loss} & {$R^{2}$} \\\midrule
DE-N-3-2-22 & 0.1056 & 0.9695 & 0.0976 & 0.9722 \\
DE-F-3-2-22 & 0.0792 & 0.9769 & 0.0663 & 0.9811 \\
DE-N-3-2-32 & 0.0887 & 0.9743 & 0.0762 & 0.9782 \\
DE-F-3-2-32 & 0.0647 & 0.9815 & 0.0692 & 0.9803 \\
DE-N-3-2-42 & 0.0704 & 0.9799 & 0.1227 & 0.9650 \\
DE-F-3-2-42 & 0.0665 & 0.9807 & 0.0685 & 0.9805 \\
DE-N-5-4-35 & 3.3562 & 0.0878 & 3.3896 & 0.0878 \\
DE-F-5-4-35 & 3.3421 & 0.0870 & 3.5537 & 0.0435 \\
DE-N-6-4-35 & 3.2840 & 0.0575 & 3.2181 & 0.0598 \\
DE-F-6-4-35 & 3.2289 & 0.0668 & 3.2293 & 0.0565 \\
\midrule
RG-N-3-2-22 & 0.0021 & 0.9994 & 0.0022 & 0.9994 \\
RG-F-3-2-22 & 0.0024 & 0.9993 & 0.0024 & 0.9993 \\
RG-N-3-2-32 & 0.0008 & 0.9997 & 0.0008 & 0.9997 \\
RG-F-3-2-32 & 0.0008 & 0.9997 & 0.0008 & 0.9997 \\
RG-N-3-2-42 & 0.0007 & 0.9997 & 0.0007 & 0.9997 \\
RG-F-3-2-42 & 0.0005 & 0.9998 & 0.0005 & 0.9998 \\
RG-N-5-4-35 & 0.0077 & 0.9979 & 0.0076 & 0.9979 \\
RG-F-5-4-35 & 0.0108 & 0.9971 & 0.0108 & 0.9971 \\
RG-N-6-4-35 & 0.0082 & 0.9976 & 0.0234 & 0.9932 \\
RG-F-6-4-35 & 0.0078 & 0.9977 & 0.0105 & 0.9969 \\\bottomrule
\end{tabular}
\end{table}

\begin{table}[t]
\caption{Mean and standard deviation of position and orientation errors on validation and test set.}\label{tab:comp_error}
\centering
\scriptsize
\resizebox{\columnwidth}{!}{\begin{tabular}{lllll}\toprule
GraphMatic & \multicolumn{2}{c}{Validation Error} &\multicolumn{2}{c}{Test Error}
\\\cmidrule(ll){2-3}\cmidrule(ll){4-5}
& {Position [cm]} & {Orientation [$^\circ$]} & {Position [cm]} & {Orientation [$^\circ$]} \\\midrule

RG-N-3-2-22 & 1.5$\pm$1.0 & 4.0$\pm$18.3 & 1.4$\pm$0.9 & 4.2$\pm$18.5 \\
RG-F-3-2-22 & 1.6$\pm$1.1 & 3.9$\pm$17.2 & 1.6$\pm$1.1 & 4.4$\pm$18.7 \\ 
RG-N-3-2-32 & 1.0$\pm$0.6 & 2.3$\pm$13.2 & 0.9$\pm$0.7 & 2.9$\pm$14.9 \\
RG-F-3-2-32 & 1.1$\pm$0.7 & 2.2$\pm$12.7 & 1.0$\pm$0.7 & 2.6$\pm$14.2 \\
RG-N-3-2-42 & 0.9$\pm$0.6 & 2.2$\pm$13.1 & 0.8$\pm$0.6 & 2.4$\pm$13.9 \\
RG-F-3-2-42 & 0.8$\pm$0.6 & 2.0$\pm$12.9 & 0.7$\pm$0.5 & 2.3$\pm$13.6 \\
RG-N-5-4-35 & 4.4$\pm$2.6 & 6.3$\pm$6.8 & 4.3$\pm$2.6 & 6.4$\pm$6.9 \\
RG-F-5-4-35 & 4.9$\pm$2.6 & 7.7$\pm$6.9 & 5.0$\pm$2.7 & 7.8$\pm$7.0 \\
RG-N-6-4-35 & 3.6$\pm$1.9 & 8.3$\pm$8.5 & 8.6$\pm$4.4& 11.4$\pm$10.4 \\
RG-F-6-4-35 & 3.3$\pm$1.8 & 8.2$\pm$8.2 & 4.5$\pm$2.6 & 8.2$\pm$7.8 \\

\bottomrule
\end{tabular}}
\end{table}

\subsection{GraphMatic-RG}
For GraphMatic-RG, the close reference angles $\boldsymbol{\tilde{\theta}}$ are randomly generated using a uniform distribution between $\pm 10^\circ$ from the target joint angle. 
Compared to GraphMatic-DE, the reference-guided variant handled the complexities of the \ik{} problem for higher \dof{}s better using the same network configurations. 
GraphMatic-RG achieved losses of $10^{-3}$ to $10^{-4}$ and $R^{2}$ values close to 1, as Tab.~\ref{learning_overview} indicates. 
The resulting position and orientation errors in $[\textrm{cm}]$ and $[^\circ]$ respectively are displayed in Tab.~\ref{tab:comp_error}.
The orientation standard deviation of 3~\dof{} is considerable, because of a specific manipulator configuration, which cannot be handled by the \gnn{}. We analyse this situation further in Section~\ref{sec:rg_errors}. Besides that, the position and orientation errors increase with increasing \dof{}, which is expected because of the increasing complexity of the \ik{}.
For 3 and 5 \dof, the test and validation results are almost the same, so we conclude the \gnn{} to be able to extrapolate unseen and uncontained link length from the training data. However, for 6 \dof, the extrapolation performed less successfully, which is indicated by a higher test error compared to the validation error. In general, we can infer that the reference-guided approach is suitable for learning an \ik{} model for more complex manipulators with higher \dof.

\subsection{Investigation of errors in GraphMatic-RG}
\label{sec:rg_errors}

\begin{figure}
     \centering
     \begin{subfigure}[b]{0.49\linewidth}
         \centering
         \includegraphics[height=0.9\textwidth]{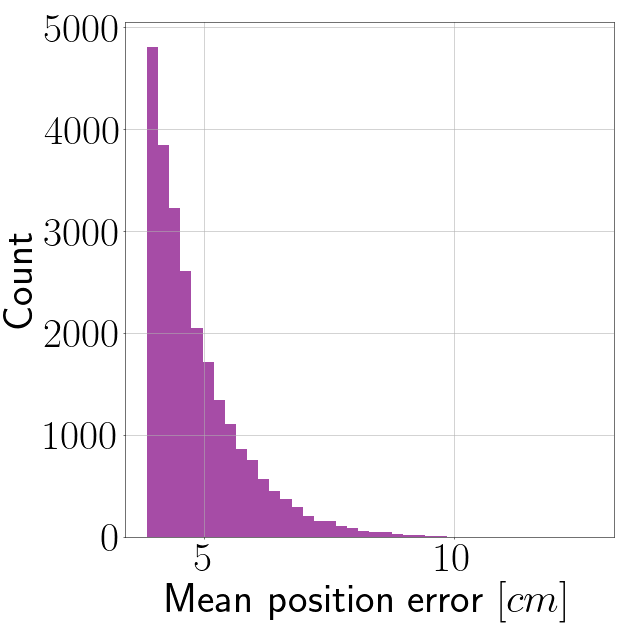}
         \caption{Position error distribution}
         \label{err_dist_pos_3}
     \end{subfigure}
     \hfill
     \begin{subfigure}[b]{0.49\linewidth}
         \centering
         \includegraphics[height=0.9\textwidth]{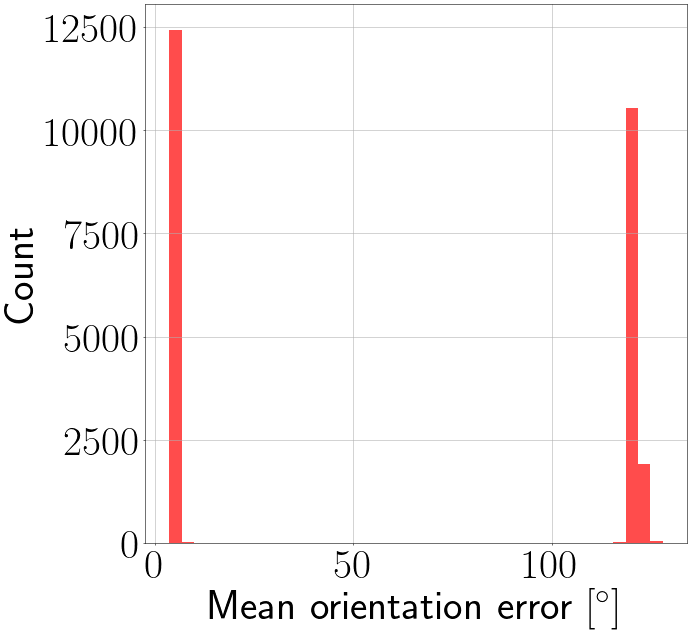}
         \caption{Orientation error distribution}
         \label{err_dist_orn_3}
     \end{subfigure}
        \caption{Error distribution of the $5\%$ test samples with the highest error of RG-F-3-2-22.}
\label{Analysis of GraphMatic-RG-F-3-2-22}
\end{figure}

\begin{figure}
     \centering
     \begin{subfigure}[b]{0.49\linewidth}
         \centering
         \includegraphics[width=\textwidth]{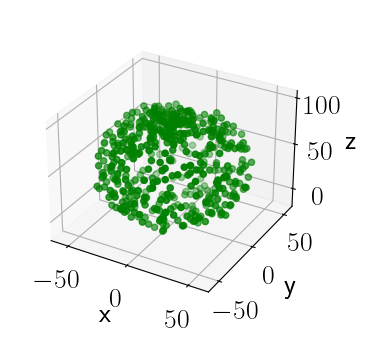}
         \caption{Position error}
         \label{pos_error5}
     \end{subfigure}
     \hfill
     \begin{subfigure}[b]{0.49\linewidth}
         \centering
         \includegraphics[width=\textwidth]{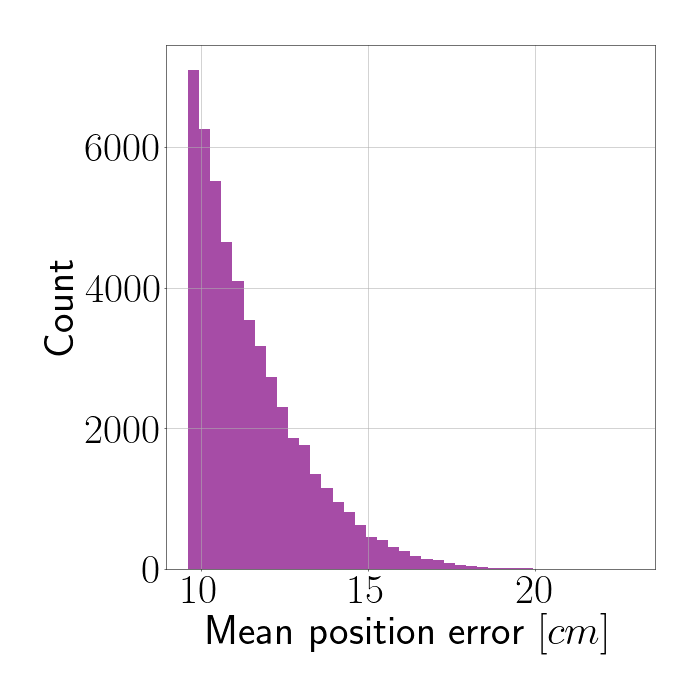}
         \caption{Position error distribution}
         \label{err_dist_pos5}
     \end{subfigure}
     \hfill
     \begin{subfigure}[b]{0.49\linewidth}
         \centering
         \includegraphics[width=\textwidth]{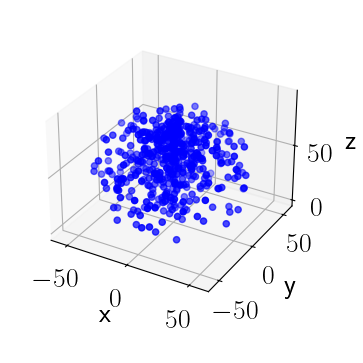}
         \caption{Orientation error} 
         \label{orn_error5}
         \end{subfigure}
         \hfill
     \begin{subfigure}[b]{0.49\linewidth}
         \centering
         \includegraphics[width=\textwidth]{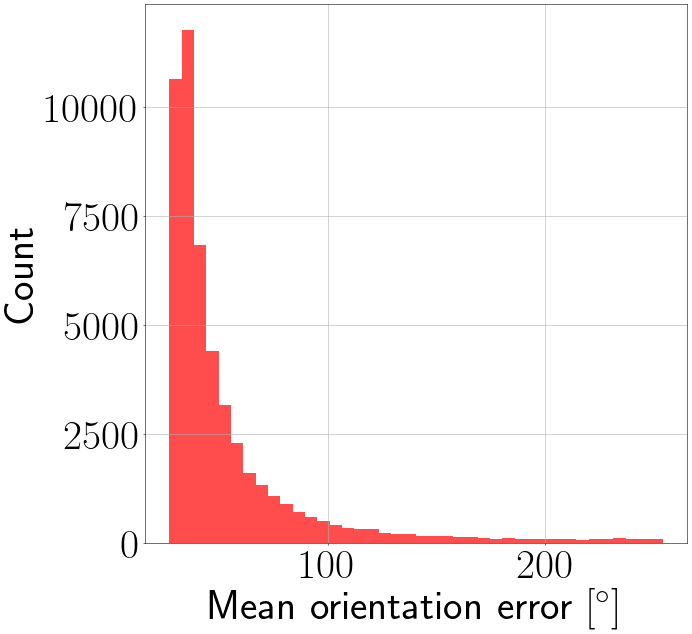}
         \caption{Orientation error distribution}
         \label{err_dist_orn5}
     \end{subfigure}
        \caption{Error distribution of the $5\%$ test samples with the highest error of RG-N-5-4-35.}
\label{Analysis of GraphMatic-RG-N-5-4-35}
\end{figure}

\begin{figure}
    \centering
    \begin{subfigure}[b]{0.49\linewidth}
        \centering
        \includegraphics[width=\textwidth]{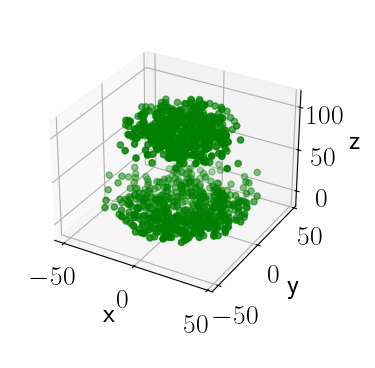}
        \caption{RG-N-6-4-35 with position error of 8.6$\pm$4.4}
        \label{GraphMatic-RG-N-test}
    \end{subfigure}
    \hfill
    \begin{subfigure}[b]{0.49\linewidth}
         \centering
         \includegraphics[width=\textwidth]{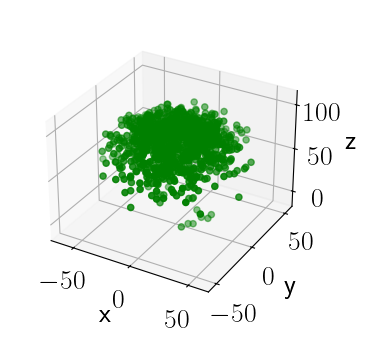}
        \caption{RG-F-6-4-35 with position error of 4.5$\pm$2.6}
         \label{GraphMatic-RG-F-test}
    \end{subfigure}
\caption{Distribution of the $5\%$ largest end-effector position errors for 6 \dof{} RG on the test set.}
\label{fig:test_dof_6}
\end{figure}

\begin{figure}
    \centering
    \begin{subfigure}[b]{0.49\linewidth}
        \centering
        \includegraphics[width=\textwidth]{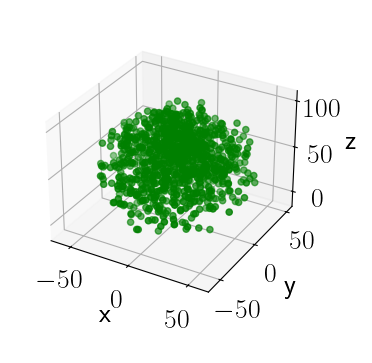}
        \caption{RG-N-6-4-35 with position error of 3.6$\pm$1.9}
        \label{GraphMatic-RG-N-val}
    \end{subfigure}
    \hfill
    \begin{subfigure}[b]{0.49\linewidth}
        \centering
        \includegraphics[width=\textwidth]{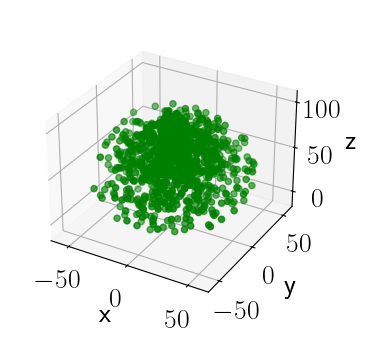}
        \caption{RG-F-6-4-35 with position error of 3.3$\pm$1.8}
        \label{GraphMatic-RG-F-val}
    \end{subfigure}
\caption{Error distribution of end-effector position for 6 \dof{} RG on the entire validation set.}
\label{fig:val_dof_6}
\end{figure}

To gain more profound insights into the prediction errors caused by GraphMatic-RG, we analyse the $5\%$ samples with the largest error on the test datasets. 
The scatter plots in Figs.~\ref{Analysis of GraphMatic-RG-N-5-4-35},~\ref{fig:test_dof_6} and~\ref{fig:val_dof_6} display the actual target positions for the $5\%$ samples with the largest error.
To avoid visual clutter, we plot every 100th sample, but the observations remain valid when all samples are plotted.
The histograms in Figs.~\ref{Analysis of GraphMatic-RG-F-3-2-22} and \ref{Analysis of GraphMatic-RG-N-5-4-35} contain all $5\%$ samples with the largest error on the test set. 
We expect the histograms to behave like a receding normal curve, which is the typical shape when training with the MSE loss function. 

\begin{figure}
    \centering
    \begin{subfigure}[b]{0.49\linewidth}
        \centering
        \includegraphics[width=\textwidth, height=0.7\textwidth]{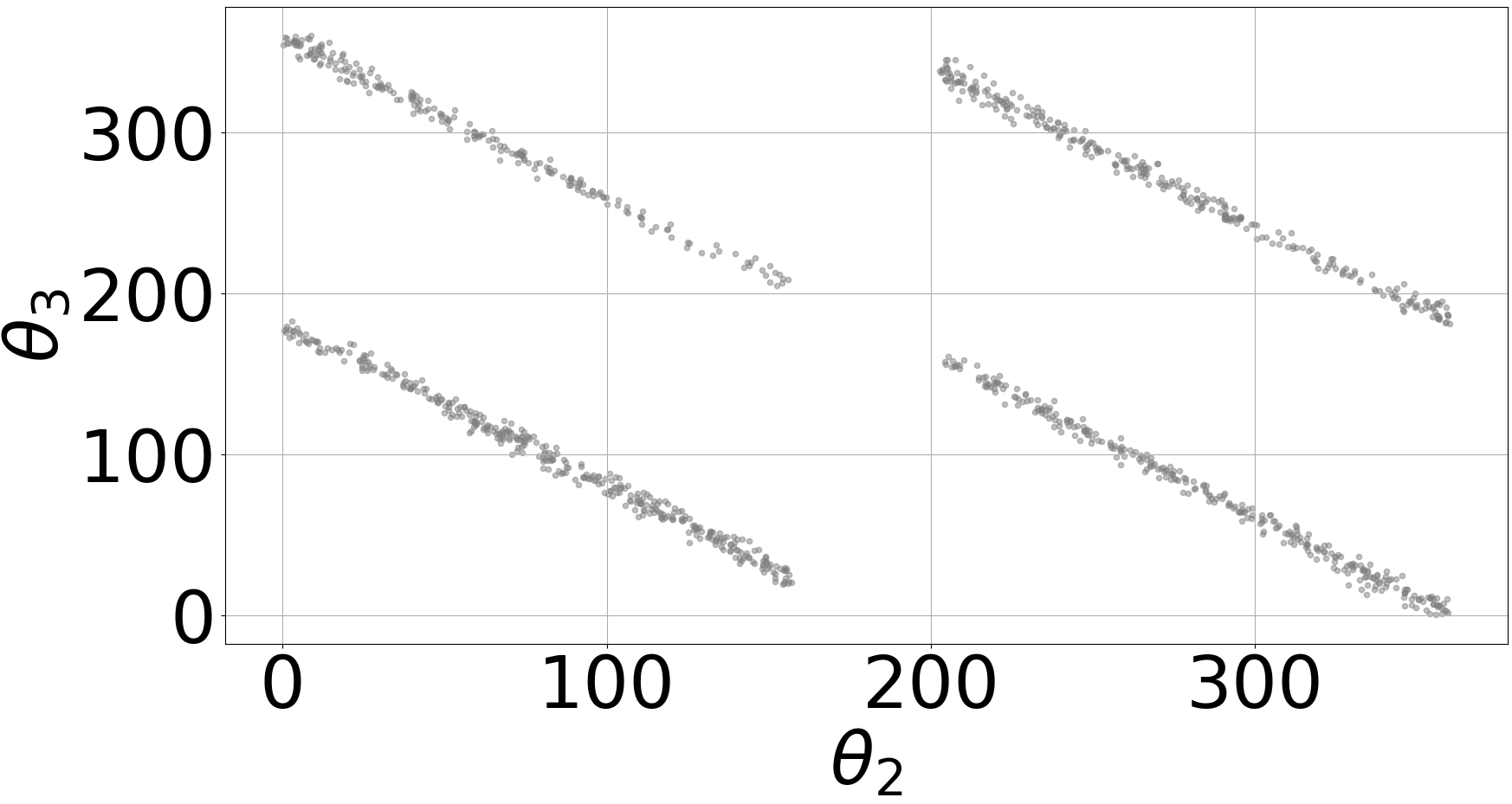}
        \caption{Expected cause of orientation error in RG-N/F for 3 \dof{} when $\theta_2 + \theta_3 \cong n\pi$ (plot axis for $\theta_2$ and $\theta_3$ are in degrees)}
        \label{orn_analysis}
    \end{subfigure}
    \hfill
    \begin{subfigure}[b]{0.49\linewidth}
         \centering
         \includegraphics[width=\textwidth, height=0.7\textwidth]{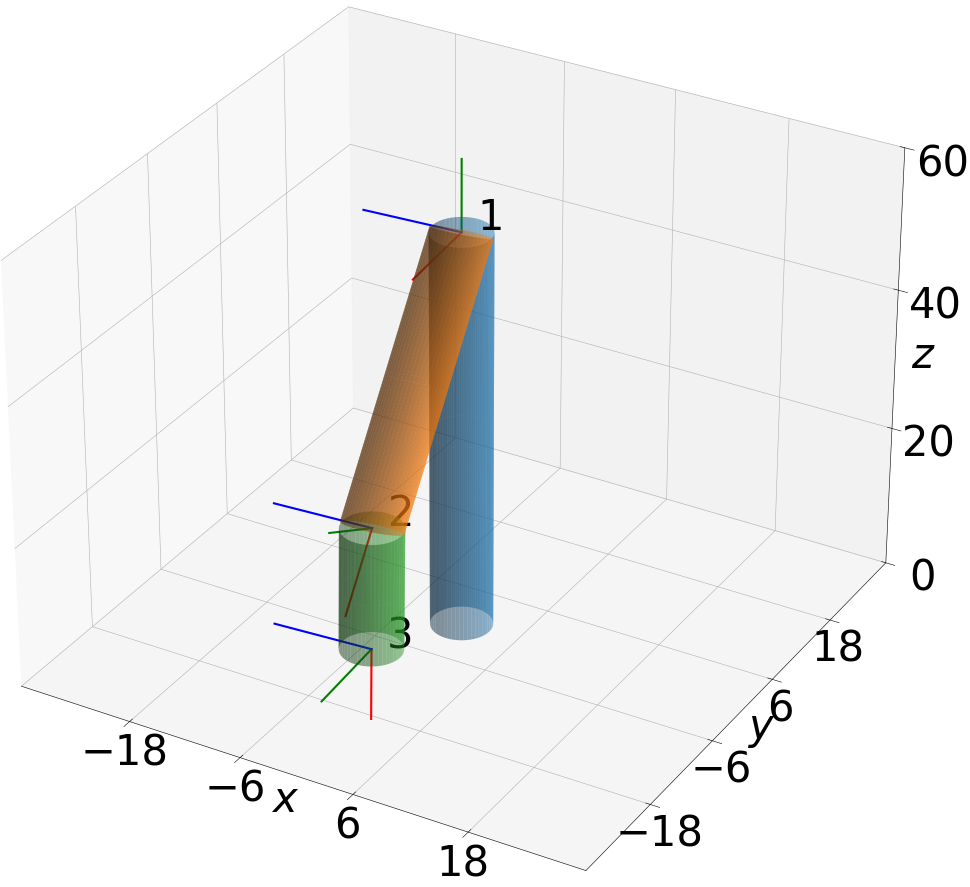}
        \caption{3 \dof{} manipulator having test set link lengths with the worst performing joint configuration, created using data generator framework}
         \label{parallel}
    \end{subfigure}
\caption{Orientation error distribution analysis for RG-F-3-2-22}
\label{nc}
\end{figure}

The analysis of RG-F-3-2-22 is shown in Fig.~\ref{Analysis of GraphMatic-RG-F-3-2-22}. 
We observe an expected behaviour for the position error distribution from Fig.~\ref{err_dist_pos_3}. 
On the other hand, the orientation error distribution in Fig.~\ref{err_dist_orn_3} is not as expected; half of the data points in consideration have a large MAE of about 120$^\circ$.
This explains the high standard deviation in 3 \dof{} orientations in Tab.~\ref{tab:comp_error}.
To analyse this behaviour, we visualize the target values for $\theta_2$ and $\theta_3$  contained in the right part of the histogram of Fig.~\ref{err_dist_orn_3}.  
We observe that the large error is triggered by the condition: $\theta_2 + \theta_3 \cong n\pi $, $n\in\{0\dots2\} \subset \mathbb{N}$. 
We visualise the 3 \dof{} manipulator with the worst orientation error in Fig.~\ref{parallel}, which shows that the last joint is facing downward and almost parallel to the z-axis. Consequently, this specific configuration seems to trigger a large orientation error. However, we could not identify the underlying reason for this behaviour, which is observed for all the network configurations for the GraphMatic-RG for 3 \dof{} manipulators.

The RG-N-5-4-35 and RG-F-5-4-35 have the expected scatter plots and error distribution for position and orientation. 
The GraphMatic with this particular configuration was able to find a solution for the \ik{}. 
The particular network configuration was able to extrapolate the link lengths, and the observed error is only where the physical constraints come into play.
In Fig.~\ref{Analysis of GraphMatic-RG-N-5-4-35}, we see a RG-N-5-4-35, but a similar behaviour is observed for the RG-F-5-4-35 as well.

Fig.~\ref{GraphMatic-RG-N-test} and Fig.~\ref{GraphMatic-RG-F-test} show the end-effector position analysis of the extreme $5\%$ worst-performing joint configurations for RG-N-6-4-35 and RG-F-6-4-35, respectively. 
We observed the expected error distributions for the pose, and the scatter plots of orientation also had the expected behaviour. 
On the contrary, the position scatter plots for RG-N-6-4-35 and RG-F-6-4-35 have unanticipated behaviour. 
RG-N-6-4-35 shows an hourglass structure of the worst performing joint configuration in Fig.~\ref{GraphMatic-RG-N-test}, whereas, for RG-F-6-4-35, the error configurations are located at the extreme top in Fig.~\ref{GraphMatic-RG-F-test}.
A possible explanation for this behaviour is that the network could not extrapolate the link lengths, because the validation set not requiring extrapolation shows the expected uniform distribution, as seen in Fig.~\ref{fig:val_dof_6}.

\subsection{Discussion}
An essential feature of a study on \ik{} models is to result in line with previous studies.
As for related works using \gnn{}s, there are no error values reported that could serve as a baseline for our approach. 
A few papers use \ann-based \ik{} models, which also report error values.
The \ann-based approach by~\cite{wagaaAnalyticalDeepLearning2023} yields a very high accuracy but also uses huge network sizes with millions of trainable parameters and comparably small dataset size. 
Their network is prone to overfitting, and the validation set contains, in the worst case, only 50 samples, which is not representative of covering the entire workspace.

The results reported in \cite{el-sherbinyComparativeStudySoft2018} are comparable to the ones presented in this paper, with 0.0016m position MSE, which we roughly translate to 4cm mean position error. 
However, they only used one manipulator configuration with 5 \dof, and do not provide a generalizing model. 
New manipulator configurations would require the model to be trained again. 

Overall, it is difficult to compare the results between works proposing ANNs to solve the \ik{} problem. 
In our paper, we put significant effort into generating trustworthy training and evaluation datasets with up to 20 million randomly generated data points for 6 \dof{}. 
Furthermore, we address a considerably more complex problem compared to other works, as we aim at learning a model that generalizes for varying link length configurations within the same manipulator family. 
Additionally, we do not limit the movement range of individual joints within the manipulators, which increases the parameter space for the \ik, and consequently, elevates the difficulty of the problem.
Finally, we also test the developed models on configurations that exceed the workspace seen during training.
Testing extrapolation capabilities is an essential step to validate \ann{} models, which we have not observed in any other study before. 

\section{Conclusion and Future Work}

In this paper, we showed \gnn{}s to be a viable option to learn the \ik{} of manipulators. They serve especially well to generalize the \ik{} of families of manipulators with the same number and type of \dof{} but variable link length. The results are promising and allow an application in certain real-world scenarios. Training data can be generated fully autonomously using the \fk{} and a simple collision check system. However, we also observed the possibility of out-of-domain errors if link length configurations were used that were not covered by the training set. This indicates the lack of extrapolation capability of \ann{}s. We hope to mitigate these problems and further increase accuracy by evaluating more parameters configurations for the \gnn{}s.
Additionally, we want to analyze the stochastic properties of the training process to gain an understanding of the reliability of the \gnn-generation. This is important, as our ultimate goal is the replacements of the \ann{} blocks in the \gnn{} by \sr{}-generated analytical equations. We envision this to solve the out-of-domain errors and the training artifacts and allow a rigorous analysis of the properties of the \ik{} enabling safety and reliability of critical real-world scenarios.

\newpage
\bibliographystyle{IEEEtran}
\bibliography{pravin_bib,christoph}
\end{document}